\documentclass[runningheads]{llncs}
\usepackage{graphicx}
\usepackage[utf8]{inputenc}
\usepackage{amsmath}
\usepackage{amssymb}
\usepackage[rightcaption]{sidecap}
\usepackage[section]{placeins}
\usepackage{color}
\usepackage{gensymb}
\usepackage{hyperref}
\usepackage{enumitem}
\newcommand{\norm}[1]{\left\lVert#1\right\rVert}
\usepackage{mathrsfs}
\usepackage{hyperref}
\usepackage{subcaption}
\usepackage{cite}
\usepackage{bm}
\usepackage[width=122mm,left=12mm,paperwidth=146mm,height=193mm,top=12mm,paperheight=217mm]{geometry}

\pdfoutput=1

\newcommand\bb{{\bm {\beta}}}

\begin{document}
\pagestyle{headings}
\mainmatter
\def\ECCVSubNumber{7381}  

\title{PnP-Net: A hybrid Perspective-n-Point Network} 

\titlerunning{ECCV-20 submission ID \ECCVSubNumber} 
\authorrunning{ECCV-20 submission ID \ECCVSubNumber} 
\author{Roy Sheffer$^{1}$, Ami Wiesel$^{1,2}$}
\institute{(1) The Hebrew University of Jerusalem (2) Google Research}
\maketitle
\begin{abstract}
We consider the robust Perspective-n-Point (PnP) problem using a hybrid approach that combines deep learning with model based algorithms. PnP is the problem of estimating the pose of a calibrated camera given a set of 3D points in the world and their corresponding 2D projections in the image. In its more challenging robust version, some of the correspondences may be mismatched and must be efficiently discarded. Classical solutions address PnP via iterative robust non-linear least squares method that exploit the problem's geometry but are either inaccurate or computationally intensive.  In contrast, we propose to combine a deep learning initial phase followed by a model-based fine tuning phase. This hybrid approach, denoted by PnP-Net, succeeds in estimating the unknown pose parameters under correspondence errors and noise, with low and fixed computational complexity requirements. We demonstrate its advantages on both synthetic data and real world data.
\end{abstract}

\section{Introduction}

\begin{figure}
\centering
\hspace{10mm}\includegraphics[width=1\textwidth]{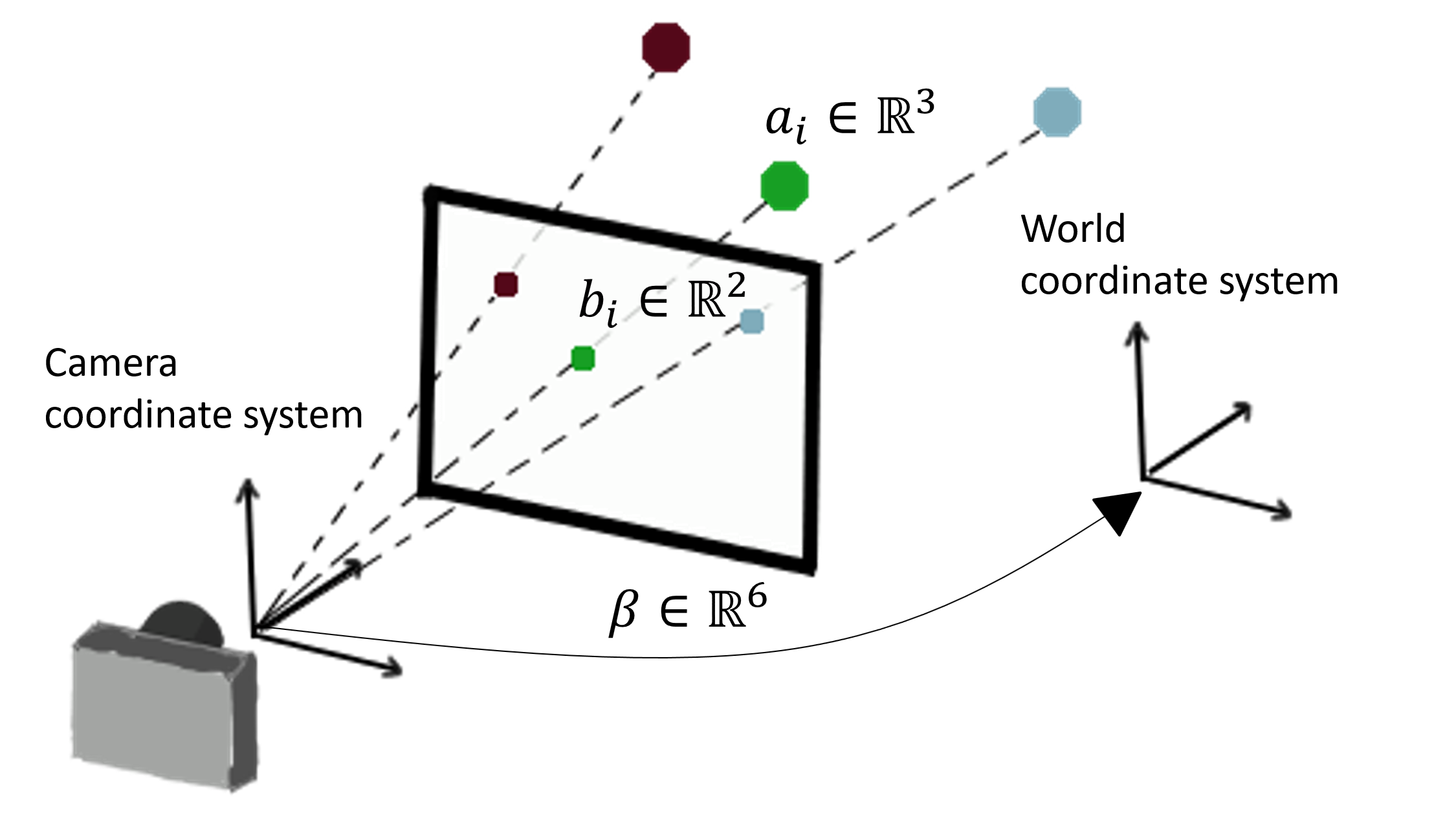}
\caption{Perspective-n-Point setting}
\label{fig:pnp}
\end{figure}

Camera pose estimation is a fundamental problem that is used in a wide variety of applications such as autonomous driving and augmented reality. The goal of perspective-n-point (PnP) is to estimate the pose parameters of a calibrated camera given the 3D coordinates of a set of objects in the world and their corresponding 2D coordinates in an image that was taken (see Fig.~\ref{fig:pnp}). Currently, PnP can be efficiently solved under 
both accurate and low noise conditions \cite{lepetit2009epnp, lm}. In practice, there may be wrong data associations within the 2D and 3D pairs due to missing objects or incorrect object recognition. Therefore, the main challenge is to develop robust PnP algorithms that are resilient to these imperfections. Identifying outliers is a computationally intensive task which is impractical using classical model based solutions. The goal of this paper is therefore to derive a hybrid approach based on deep learning and classical methods for efficient and robust PnP.



The PnP geometry gives rise to a well defined non-linear least squares (LS) formulation. The latter is traditionally solved using the well known Levenberg-Marquardt (LM) method \cite{lm}. Robust variants of LS can be derived via iteratively re-weighted least squares (IRLS) \cite{irls}. These minimize the squared re-projection errors of the given inputs, while weighting each sample inversely to its re-projection error as induced by the current parameters. Due to the non-linearities and weight functions, such methods are highly non-convex and are very sensitive to their initialization. Consequently, real systems must resort to computationally intensive search methods. The popular RANSAC \cite{ransac} algorithm repeats the LM methods on different subsets and matching possibilities, ultimately choosing the pose with the smallest squared re-projection errors. RANSAC is an efficient randomized algorithm, yet it is still too  computationally intensive for real time deployment.
Recently, deep learning has been shown to be highly successful at image recognition tasks \cite{resnet, yolo}, as well as other domains including geometry
\cite{deppdepths, deepMatch}. There is a growing body of literature on unfolding iterative algorithms into deep neural networks \cite{gregor2010learning,hershey2014deep,samuel2017deep}. Part of these recent successes is the ability to perform end-to-end training, i.e. propagating gradients back through an entire pipeline to allow the direct optimization of a task-specific loss function, e.g., \cite{thewlis2016fully, lift}. 
Motivated by these ideas, we propose a hybrid deep learning and IRLS-LM approach to robust PnP. Experiments with deep learning on its own, suggest that accurate PnP estimation is too difficult for existing neural networks. Instead, we take a different route and use the network to robustly compute a coarse pose estimation. We then use it as initialization to an IRLS-LM solver and then jointly optimize both stages simultaneously in an end-to-end fashion. We train the system using synthetic data generated according to a theoretical PnP model.
The overall solution, denoted by PnP-Net, achieves a flexible accuracy vs complexity tradeoff and outperforms state of the art methods on both its true and mismatched models. Remarkably, even though PnP-Net was trained on synthetic data, numerous experiments in different scenarios including real world data show accuracy similar to RANSAC with computational complexity on the order of classical LM.

\subsection{Related Work}\label{sec:related}
PnP is a fundamental problem in computer vision and has a long history of parameter estimation methods. We refer the reader to \cite{pnpreview} for a comprehensive review. Below, we briefly review the leading algorithms along with their pros and cons.

The classical solution P3P addresses the case of $n=3$ input correspondences \cite{gao2003complete}. Its main drawback is that it is sensitive to noise. An efficient extension to more points is  EPnP \cite{lepetit2009epnp}. Using more points reduces the sensitivity to noise and provides better accuracy. The algorithm ignores the nonlinear constraints and uses linearization to express the multiple points as a weighted sum of four virtual control points. Similarly to P3P, it too is sensitive to noise. REPPnP provides a more robust version that rejects outliers \cite{reppnp}. REPPnP requires a large number of correspondences and obtains poor results when only a small number of correspondences is available.

The direct approach to outlier rejection is via a brute force search over the correspondences mismatches. Typically, a stochastic approximation known as RANSAC samples random sets of correspondences and computes their estimates \cite{ransac}. Its model hypotheses can be any of the previously mentioned algorithms, e.g., RANSAC-with-P3P or RANSAC-with-EPnP. Unfortunately, the computational complexity of both the full and approximate searches is intractable for real-time practical systems.

From an optimization perspective, PnP is a nonlinear LS minimization. The classical approach to its solution is LM which can be interpreted as a regularized Gauss Newton method \cite{lm}. LM is a scalable solution that exploits the underlying structure of the PnP problem. The squared loss is sensitive to outliers and fails when there are correspondence mismatches in the inputs. It is well known that LS methods can be robustified by replacing the squared loss with robust loss functions, e.g., Huber's loss, and iteratively re-weighting the algorithms \cite{irls}. Together, the combination of IRLS with LM is a natural and promising approach to robust PnP. But, as a non-convex optimization, it requires good initialization strategies to avoid spurious local minima.


More recently, there is a growing body of works using deep learning.
The preprocessing stage of PnP-Net is based on the ideas in
PointNet for classification on a cloud of points  \cite{pointnet}. 
Similarly,  \cite{deep_correspondences} uses deep learning for finding robust image to image correspondences, but does not consider pose estimation. 
Perspective-n-Learned-Point \cite{PerspectivenLearnedPoint}  combines
Content-Based Image Retrieval and pose refinement based on a learned representation of the scene geometry extracted from monocular images. 
Probably closest to our work, BA-Net \cite{BaNet} introduces a network architecture to solve the structure-from-motion (SfM) problem via feature-metric bundle adjustment (BA), which explicitly enforces multi-view geometry constraints in the form of feature-metric error. Like PnP-Net, this work also relies on a differentiable pipeline that includes an LM step. On the other hand, the last two networks consider image inputs whereas PnP-Net 
follows the more common practice used in real systems which is based on 3D coordinate inputs. Finally, to our knowledge, PnP-Net is the first work to use a neural network as an initialization to an unfolded algorithm with end to end training. This is the unique contribution of our work and we believe it will be useful in other problems as well. 

\subsection{Notation}
We denote the elements of a 3D vector as $[a_x,a_y,a_z]$.
The perspective function ${\bm \Pi}:  \mathbb{R}^3 \rightarrow  \mathbb{R}^2$ is defined as ${\bm \Pi}({\bf a})=\left[\;a_x/a_z,\;a_y/a_z\;\right]$.
Following \cite{Rodrigues}, we represent a 3D rotation via a 
 rotation angle $\theta \in[0, \pi]$ and a normalized rotation axis according to the right hand rule ${\bf s}\in \mathbb{R}^3$ where $\norm{{\bf s}} = 1$. The rotation matrix ${{\bf R}}$ can then be constructed as:
\begin{equation} \label{eq:R}
{\bf R}(\theta,{\bf{s}}) = {\bf I} + \sin(\theta){\bf M} + (1-\cos(\theta)){\bf M}^2
\end{equation}
\begin{equation} \label{eq:M}
{\bf M}=
\begin{pmatrix}
0&-s_{z}&s_{y}\\
s_{z}&0&-s_{x}\\
-s_{y}&s_{x}&0\\
\end{pmatrix}
\end{equation}
We measure the difference between two rotations using the metric
\begin{equation}
    d({{\bf R}}_1,{{\bf R}}_2)=\arccos{\frac{{\rm{Tr}}\left[{{\bf R}}_1{{\bf R}}_2^T\right]-1}{2}}.
\end{equation}


\section{Problem Formulation}
In this section, we define the PnP model and formulate the parameter estimation problem. The unknown rigid transformation parameters are the pose of a camera, defined by its translation parameters ${\bf t}$, and Rodrigues rotation parameters $\theta$ and ${\bf s}$. 
Together, we denote these parameters as $\bb = \{ {\bf t}, \theta,{\bf s}\}$.

The PnP model provides access to $n$ pairs of 3D objects in world coordinate system, denoted by ${\bf a}_i$, and their corresponding 2D coordinates on the image plain denoted by ${\bf b}_i$. Ideally, these pairs satisfy a projection transformation parameterized by $\bb$:
 \begin{eqnarray}\label{transf}
     {\bf b}_i &=& C_\bb({\bf a}_i)  \nonumber\\
 &=& {\bm \Pi}\left({\bf K} \left[{\bf Ra}_i + {\bf t}\right]\right) \qquad i=1,\cdots,n.
 \end{eqnarray}
 where ${\bf{K}}$ is the camera intrinsic matrix \cite{hartley}:
\begin{equation} \label{intrinsic:1}
{\bf K} =
\begin{pmatrix}
f&0&0\\
0&f&0\\
0&0&1\\
\end{pmatrix}
\end{equation}
and $f$ is the known focal length of the camera.


In practice, the transformation in (\ref{transf}) is inexact. First, both sides of the equation are corrupted by additive noise. Second, there may be mismatches between some of the 3D points and their 2D projections, i.e.,  the pairs $\{{\bf a}_i,{\bf b}_i\}$ and $\{{\bf a}_j,{\bf b}_j\}$  for some $i\neq j$ may be mistaken as the pairs $\{{\bf a}_i,{\bf b}_j\}$ and $\{{\bf a}_j,{\bf b}_i\}$. 

The robust PnP problem is to recover the unknown $\bb$ given the inexact pairs $\{{\bf a}_i,{\bf b}_i\}$ for $i=1,\cdots,n$. Performance is measured using a Euclidean translation error
\begin{eqnarray}\label{distt}
\epsilon_t = \left\|\hat{\bf{t}}-{\bf{t}}\right\|
\end{eqnarray}
and the rotation error
\begin{eqnarray}\label{distr}
\epsilon_r = d({\bf{R}}(\hat\theta,\hat{\bf{s}}),{\bf{R}}(\theta,{\bf{s}}))
\end{eqnarray}
where $\hat{\bf{t}}$, $\hat\theta$ and $\hat{\bf{s}}$ are the estimated parameters.

\section{PnP-Net}
\begin{figure} [!h] 
\centering
\includegraphics[width=1\textwidth]{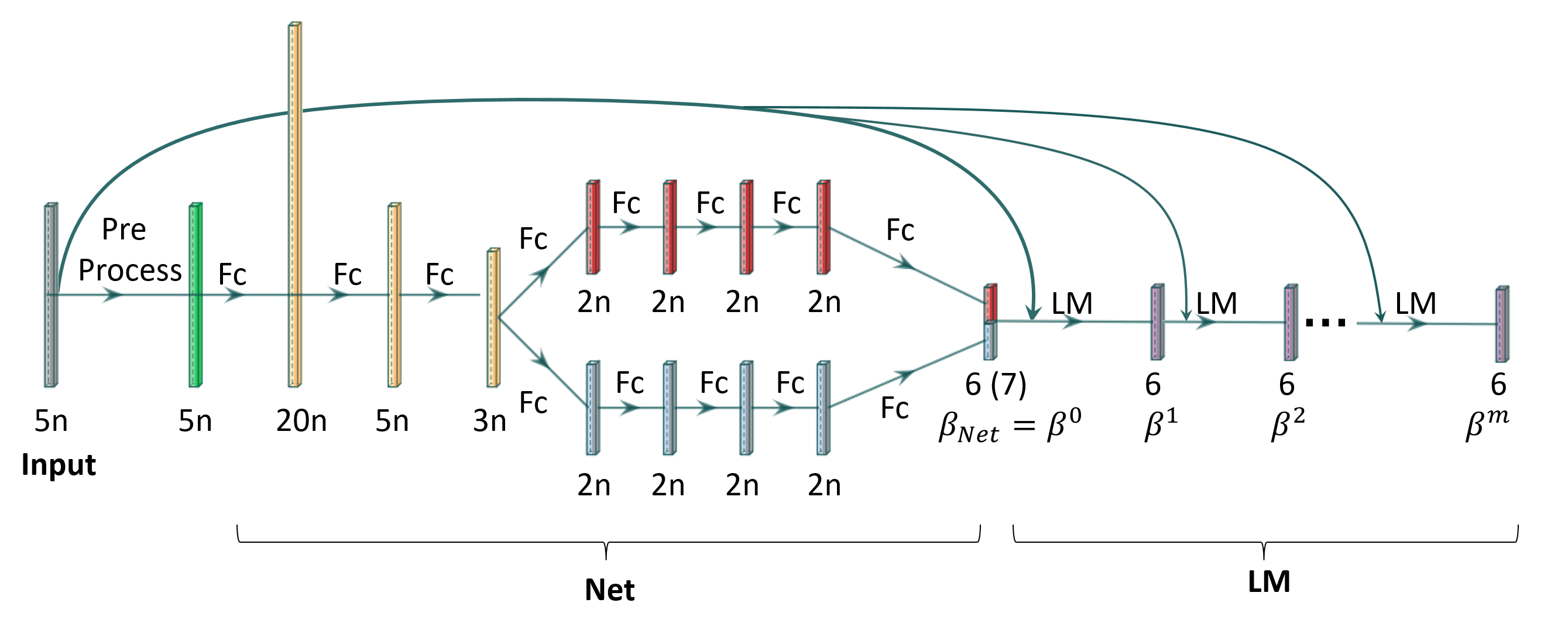}
\caption{PnP-Net diagram.}
\label{fig:diagram}
\end{figure}

In this section, we propose PnP-Net a hybrid solution that combines the model-based advantages of classical PnP methods with the power of modern deep learning. The main idea is that robust LM methods exploit the structure of the problem and are near optimal when given a sufficiently accurate initialization. On the other hand, neural networks are ignorant of the geometry and the underlying model, but are useful in learning a coarse initialization. Therefore, PnP-Net begins with a general purpose neural network which is robust and sufficiently accurate to guide the near-optimal model-based IRLS-LM phase. We learn both stages in an end-to-end manner by unfolding the iterative phase as part of the network. This allows us to learn data-driven hyper-parameters, including step sizes, regularization parameters, and weight functions. PnP-Net is fitted using synthetic data generated according to the the PnP parametric model.
The result is a low and fixed computational complexity algorithm with state-of-the-art accuracy. The network is illustrated in Fig.~\ref{fig:diagram} and can be divided to three stages detailed below.

\subsection{Preprocessing}
To simplify the setting, PnP-Net begins with a standard preprocessing stage. First, the 
image coordinates are normalized to emulate a camera with a constant focal length $f_{\rm{const}}$ instead of their known $f$, i.e., ${\bf{a}}_i$ are replaced by $\frac{f_{\rm{const}}}{f}{\bf{a}}_i$.
This normalization avoids forcing the network to learn how to handle different focal lengths. Second, following \cite{pointnet}, we order the correspondence pairs by their image coordinates to enforce order invariance.

\subsection{Net} \label{net_section}
The architecture of PnP-Net begins with a general purpose neural network whose inputs are  $\{{\bf a}_i, \ {\bf b}_i\}_{i=1}^{n}$ for fixed $n$ and outputs coarse estimates of both the translation ${\bf{t}}$ and rotation parameters $\theta{\bf{s}}$. The rotation estimate is represented by $4$ numbers, $3$ unnormalized rotation axis parameters $s$ and $1$ rotation angle parameters $\theta$. The network is built of fully connected layers with ReLU activations.  

The network begins with $3$ relatively large fully connected layers of sizes $20n$, $5n$ ,$3n$ that are shared across both tasks.
Intuitively, this stage is supposed to identify the anomalous correspondences are weight them accordingly. 
Next, the network splits into two decoupled flows of $5$ smaller layers of size $2n$. The last layer in each flow is linear and outputs coarse estimates of the rotation and translation parameters, respectively. Due to the redundancy in the rotation parameterization, the output of this stage is of dimension $7$ and not $6$.


\subsection{LM}

Given the coarse neural pose initialization, the second phase of PnP-Net refines it using a time-tested model-based approach. Robust PnP involves two challenges: outliers and non-linearity. In what follows, we review the two classical approaches to these challenges and then merge them together into a unified unfolded architecture.

The state of the art approach to real-time non-linear LS in computer vision is the LM algorithm \cite{lm,fletcher1971modified}. 
Its main idea is to solve a sequence of linearized LS problems that have a closed form solution, together with a form of Tikhonov regularization. In out setting, LM minimizes  the sum of squares
\begin{equation}
\min\limits_{\bb}\ \sum_{i=1}^{n}\left\|{\bf r}_i\right\|^2
\end{equation}
of re-projection errors
\begin{equation}
{\bf r}_i = C_{\bb}({\bf a}_{i}) - {\bf b}_{i}.
\end{equation}
In each iteration, it linearizes the errors around the previous estimates
\begin{eqnarray}
C_{\bb+{\bm \delta}}({\bf a}_{i}) \approx C_{\bb}({\bf a}_{i}) + {\bf J}_i{\bm \delta} 
\end{eqnarray}
where 
\begin{equation}
    {\bf J}_i = \frac{\partial C_{\bb}({\bf a}_{i})}{\partial \bb}
\end{equation}
is the gradient row-vector of $C_{\bb}({\bf a}_{i})$ with respect to $\bb$ evaluated at $\bb$. Stacking the errors and the gradients of the $n$ points, yields the residual vector  ${\bf r}\in \mathbb{R}^{2n}$ and the Jacobian ${\bf J} \in \mathbb{R}^{2n\times6}$, respectively. The update ${\bm \delta}$ can then be computed by solving a linear system
\begin{eqnarray}
({\bf J}^{T}{\bf J} + \lambda {\rm diag}({\bf J}^{T}{\bf J})){\bm \delta} = {\bf J}^{T}{\bf r}
\end{eqnarray}
where $\lambda$ is a regularization parameter. The resulting iteration is defined as
\begin{eqnarray}
\bb^{j} = \bb^{j-1} + \gamma \cdot {\bm \delta}\end{eqnarray}
where $\gamma$ is a step size parameter. This is currently the leading solution to PnP in practical systems. 


On the other hand, the classical approach to {\em robust} LS relies on robust loss functions, e.g., replacing the squared loss with Huber's loss denoted by $\rho(\cdot)$ \cite{irls}:
\begin{equation}
\min\limits_{\bb}\ \sum_{i=1}^{n}\rho\left(\|{\bf r}_i\|\right)
\end{equation}
Numerically, this optimization is traditionally minimized by solving a sequence of re-weighted LS problems (IRLS): 
\begin{equation}
\bb^{j} = \text{arg}\,\min\limits_{\bb}\ \sum_{i=1}^{n}w_{i}^{j}||C_{\beta}({\bf a}_{i}) - {\bf b}_{i}||^2
\end{equation}
where
\begin{equation}\label{weight}
 w_{i}^j = \frac{1}{\|C_{\bb^{j-1}}({\bf a}_{i}) - {\bf b}_{i}\|^{\alpha}}
\end{equation}
is the $i$'th correspondence weight in the $j$'th optimization iteration and $\alpha \in \mathbb{R}^{+} $ is a robustness parameter. For example, choosing $\alpha=1$ corresponds to an absolute deviation loss which is known to be more robust. Intuitively, this approach down weights bad correspondences and softly rejects outliers. 

As explained above, PnP involves a {\em non-linear} and {\em robust} LS. Therefore, we propose to combine these two ideas and solve a sequence of both re-weighted and linearized models. This yields the following linear system
 \begin{eqnarray}
({\bf J}^{T}{\bf W}{\bf J} + \lambda {\rm diag}({\bf J}^{T}{\bf W}{\bf J})){\bm \delta} = {\bf J}^{T}{\bf W}{\bf r}
\end{eqnarray}
where ${\bf W}\in\mathbb{R}^{2n\times2n}$ is a diagonal matrix with the weights $w_i^j$
\begin{eqnarray}
W^{j} =
  \begin{bmatrix}
    w_{0}^j & & & &\\
    & w_{0}^j & & & &\\
    & & \ddots & & &\\
    & & & w_{n}^j & \\
    & & & & w_{n}^j.
  \end{bmatrix}.
\end{eqnarray}

Altogether, we unfold $m$ of these iterations as layers, and define $\alpha$, $\gamma$ and $\lambda$ as learned variables. The inputs to each layer are the original correspondence pairs and the previous estimate, whereas the output is the new estimate. The first layer uses the neural Net initialization as its estimate. The complete architecture is illustrated in Fig. 2. All its unknown variables are learned simultaneously in an end to end manner.

\subsection{Training protocol}
PnP-Net is trained in an end to end manner including both stages using synthetic data generated from the PnP model.

\begin{itemize}
    \item {\bf{Training data:}} PnP-Net is trained using synthetic data generated from the PnP model. The translation ${\bf t}$, rotation axis {\bf s} and rotation angle $\theta$ parameters are uniformly distributed in a $25\times25\times25$ 3D box centered at the origin of the world, the unit ball and $[0, \frac{\pi}{2}]$, respectively.
In practice, this is equivalent to centering the world coordinates around a coarse estimation of the camera pose which usually can be obtained by a GPS. The 3D inputs ${\bf a}_i$ were uniformly distributed in a $20\times20\times80$ 3D box placed in front of the camera, and their 2D projections were computed via (\ref{transf}). The inputs to the network were then corrupted by additive Gaussian noises with variances $\sigma_1 = 0.05$, $\sigma_2 = 1$. Finally, the data included a uniformly distributed ratio in $[0,  \frac{n}{3}]$ of bad correspondences. In half of these outliers, we modeled wrong matching and the 2D projections were generated from the true model but wrong 3D coordinates. In the other half, we modeled wrong sensing and uniformly generated 2D coordinates within the image. A motivating setting for such outliers is a camera that is mounted on a car and the lower half of its image contains only the road without any traffic signs. On the other hand, an arbitrary object on the road might be mistakenly identified as a traffic sign and the network must experience this type of wrong matches. 


\vspace{5mm}

\item {\bf{Loss function:}} The end to end loss is a weighted sum of four distances. We introduce an intermediate loss after the Net phase and a final loss after the LM phase. Each of these is a sum of the Euclidean translation error in (\ref{distt}) and the rotation error as defined in (\ref{distr}). The LM phase is highly non-linear and much harder to optimize. Therefore, the training begins with the intermediate losses alone and gradually adds the final losses as the network converges. The exact protocol is detailed in the supplementary code.



\vspace{5mm}

\item {\bf{Implementation Details:}}
PnP-Net was trained using TensorFlow \cite{abadi2016tensorflow}, with the Adam optimizer and a learning rate of $10^{-4}$. Each mini-batch consisted of $1000$ randomly generated samples, and we trained for a total of 200k updates. More details are available in the supplementary code.

\end{itemize}


\section{Experiments}
In this section, we demonstrate the advantages of PnP-Net with respect to state of the art. Performance will be measured in terms of the tradeoff between estimation accuracy and computational complexity. Furthermore, unlike classical PnP algorithms, PnP-Net is based on machine learning and is prune to overfitting. Therefore, we will also test PnP-Net's ability to generalize to realistic problem settings. 

We examine the performance of the following competitors:
\begin{itemize}
    \item \textbf{PnP-Net}:  The complete PnP-Net algorithm ($\bb^{LM}$).
        \item \textbf{Net}: Only the Net stage of PnP-Net ($\bb^{Net}$).
\item \textbf{EPnP}:  \cite{lepetit2009epnp}.
    \item \textbf{EPnP-LM}: An EPnP initialization  \cite{lepetit2009epnp} followed by LM.
    \item \textbf{REPPnP} \cite{reppnp}.
    \item \textbf{REPPnP-LM}: An REPPnP initialization \cite{reppnp} followed by LM.
    \item \textbf{Ransac} \cite{ransac} with \textbf{EPnP-LM}  \cite{lepetit2009epnp, lm} as hypotheses generator and minimal subsets of size 7.
\end{itemize}

The algorithms with the LM suffix are variants of their prefix method with an additional refinement using our proposed IRLS LM. We are not aware of specific references on such two step approaches, but they improve all methods and we decided to add them to ensure a fair comparison. Note however that only PnP-Net jointly learns both stages. 

Practical autonomous driving systems are based on different modes of operations requiring different accuracy metrics. These are defined by threshold parameters  $t_{R}$ and  $t_{T}$ for the rotation and translation estimation errors, respectively. A rotation estimation is defined as successful if $\epsilon_r(\beta') < t_{R}$, and a translation estimation is successful if  $\epsilon_t(\beta') < t_{T}$. If both parameters are successfully estimated, then the complete pose estimation is successful. In practice, once these error thresholds are achieved, updating the pose is a much simpler task by tracking objects image movement and using \cite{egomotion}. 
In the results below, we report the percentage of successful pose estimations out of the total number of tests. 

\subsection{Computational complexity}
The main motivation to PnP-Net is to provide near optimal accuracy with reduced computational complexity. To emphasize this property we begin with a computational complexity analysis by counting the number of operations in each algorithm. 
The counting was performed following the protocol in \cite{flops} and the code is available in the supplementary material. The counts in Fig.~\ref{fig:ops} illustrate the low complexity of all the algorithms in comparison to the exponential complexity of RANSAC. As promised, the complexity of PnP-Net is also  significantly lower, and even lower than the REPPnP competitors. 

The complexity curves also decouple the cost of the two phases in PnP-Net. As expected, the neural initialization phase denoted by Net adds a negligible complexity. In what follows, we show the accuracy improvement due to this low cost initialization.

\begin{figure}[!h] 
    \centering
    \begin{minipage}{.5\textwidth}
        \centering
        \includegraphics[width=\textwidth]{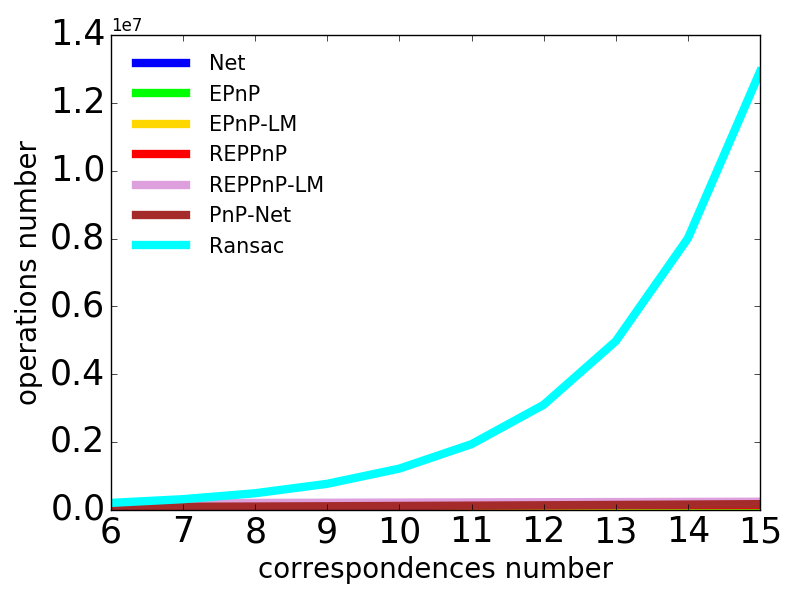}
        \label{fig:operation_count_all}
    \end{minipage}%
    \begin{minipage}{0.5\textwidth}
        \centering
        \includegraphics[width=\textwidth]{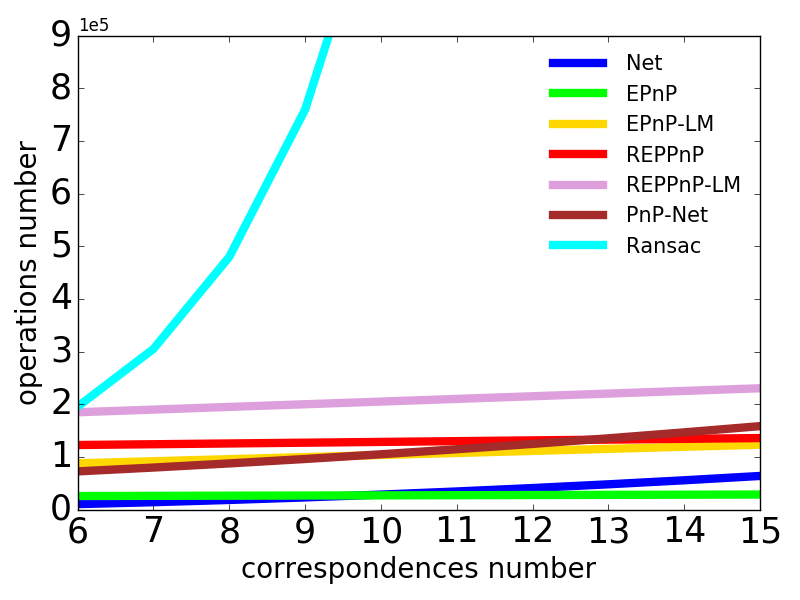}
        \label{fig:operation_count_no_ransac}
    \end{minipage}
\caption{Operation count as a function of correspondences number. Zoom out on the left, zoom in on the right.} 
\label{fig:ops}
\end{figure}

\subsection{Synthetic Data}
\begin{figure} [!h] 
\centering
\includegraphics[width=1\textwidth]{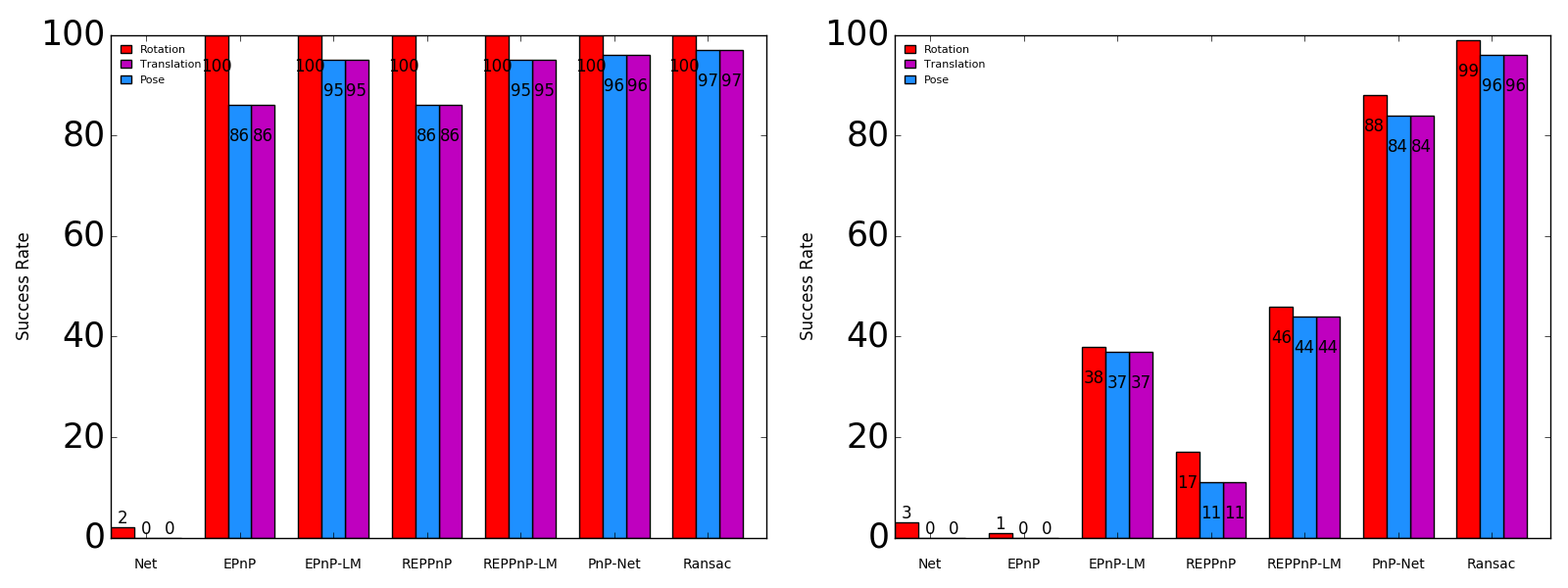}
\caption{Success rates of different algorithms on synthetic data. Near optimal input on the left and near optimal input with outliers on the right.} 
\label{fig:syn}
\end{figure}

We begin with simple experiments with synthetic data generated in the same manner as in the training process but independently. We measure our accuracy by measuring the success rate with $t_{R}=1$  $t_{T}=0.2$ over $1000$ different inputs in comparison to the other existing solutions described above. The results are shown in Fig.~\ref{fig:syn}. It is easy to see that all the algorithms provide accurate estimates in the left panel when there are no outliers. On the other hand, the right panel provides the success rates with outliers. In this more challenging scenario, all the algorithms fail expect for PnP-Net and the computationally expensive RANSAC. 

We note the low success rate of the Net stage alone in both settings with and without outliers. The reason is that the neural network fails in achieving our stringent accuracy specifications, and is only useful as a coarse initialization of PnP-Net.

\begin{figure} [!h] 
\centering
\includegraphics[width=1\textwidth]{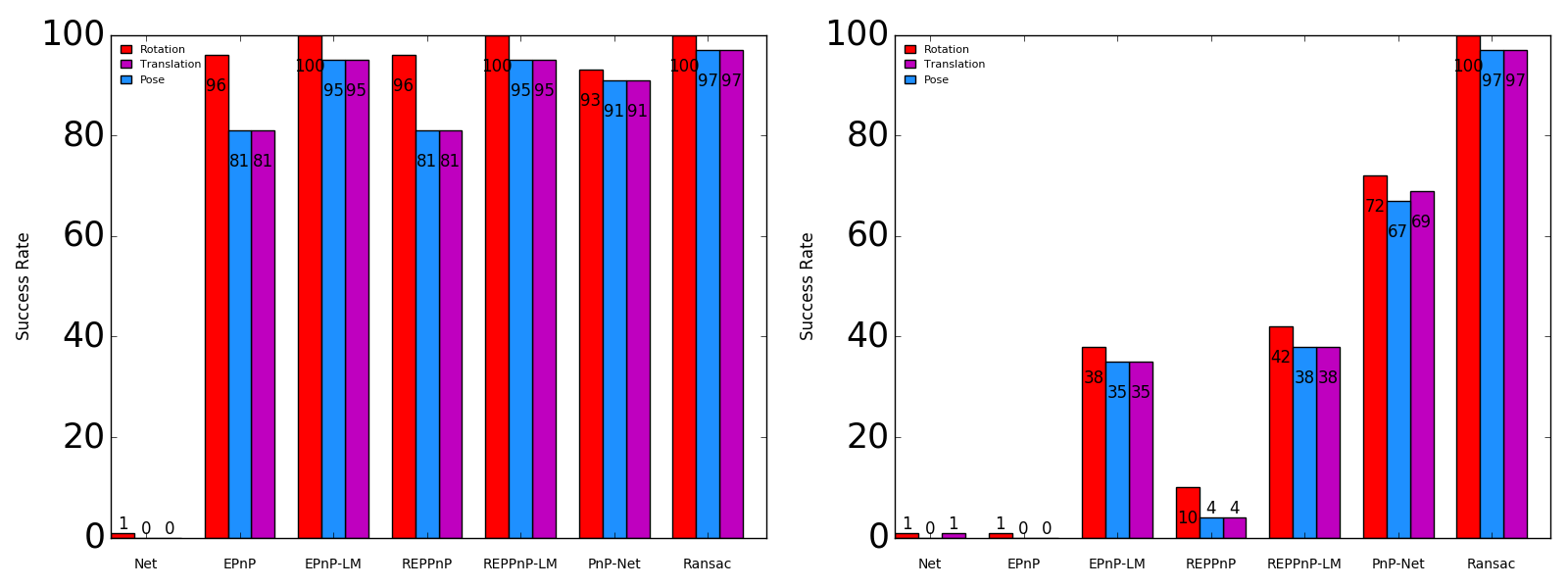}
\caption{Success rates of different algorithms on synthetic data with different random pose generation from the training. Near optimal input on the left and near optimal input with outliers on the right.} 
\label{fig:Gaussian_pose}
\end{figure}

In our second and more challenging synthetic experiment, we examine the ability of PnP-Net to generalize well to settings different than those it was trained on. For this purpose, we repeated the previous experiment but changed the prior distributions of the pose translation (${\bf t}$). We switched the uniform prior to a normal distribution with zero mean and variance $\sigma=25$ and $\mu=0$ on each of the axes. This choice allows translations vectors outside the uniform box that were never encountered during training.  As shown in Fig.~\ref{fig:Gaussian_pose}, PnP-Net's success  rates decrease but still dominate its competitorrs. Performance without outliers in the left panel reveals that all algorithms behave similarly, and PnP-Net does not overfit. The main advantage is again the robustness to mismatches as illustrates in the right panel. PnP-Net is significantly more accurate than all its competitors except the computationally expensive RANSAC.


\subsection{Real Data}
\begin{figure} [!h] 
\centering
\hspace{10mm}\includegraphics[width=1\textwidth]{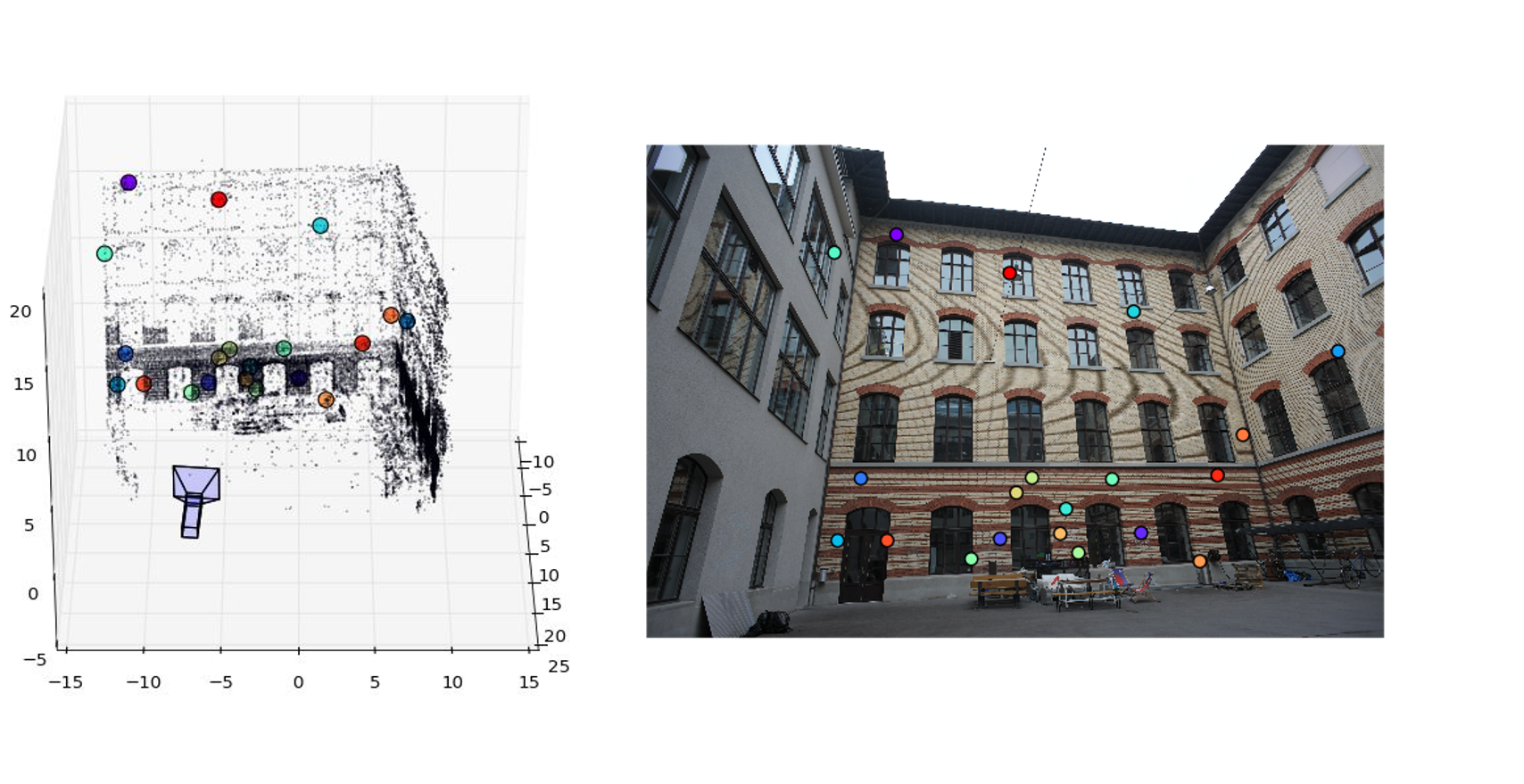}
\caption{ETH3D Benchmark data-set objects 3D coordinates and their image correspondences marked by the same color.}
\label{fig:dataset}
\end{figure}

Motivated by the synthetic results we turn to a real data-set from the ETH3D Benchmark \cite{eth3d}. The data-set contains a set of objects 3D coordinates, images in which these objects can be seen, the intrinsic parameters and the pose of each of the cameras with respect to the same coordinate system as the objects 3D coordinates. Examples of objects and their respective coordinates in the image are provided in Fig.~\ref{fig:dataset}. 

\begin{figure} [!h] 
\centering
\includegraphics[width=1\textwidth]{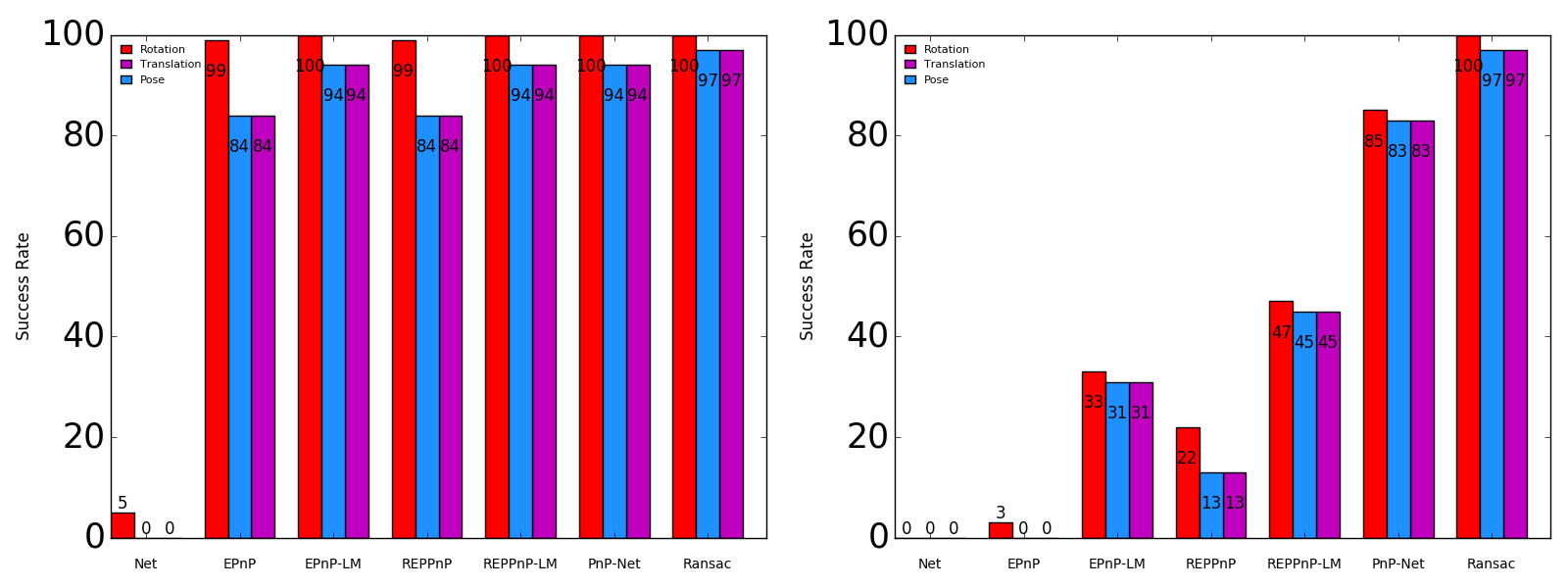}
\caption{Success rates of different algorithms on real data. Near optimal input on the left and near optimal input with outliers on the right.} 
\label{fig:real}
\end{figure}

The evaluation was done using $t_{R}=1$,  $t_{T}=0.2$ and $1000$ inputs as in the synthetic data, generated by randomly choosing an image out of the data set \cite{eth3d}, then randomly choosing 9 objects from the image  that together with their image coordinates and the focal length were given as input to each of the algorithms. For the inputs with outliers, $10$ objects were chosen, then $2$ of the first $9$ chosen objects were randomly chosen as outliers. 2D coordinates outliers were replaced by either the $10$'th unused  object image coordinates or by uniformly sampled 2D image coordinates. The results with and without mismatches are provided in Fig.~\ref{fig:real} and are similar to the synthetic experiments. This strengthens our learning approach with synthetic data and suggests that it generalizes well to realistic conditions without overfitting.

\begin{figure} [!h] 
\centering
\includegraphics[width=1\textwidth]{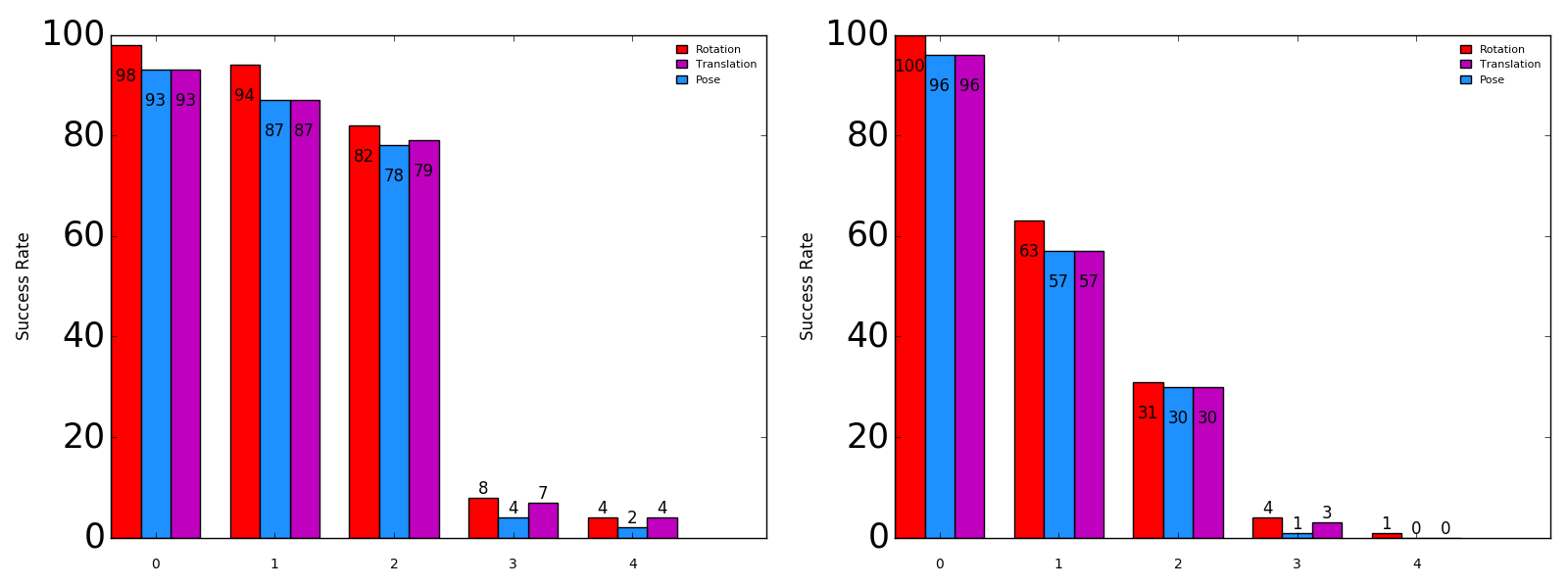}
\caption{Success rates of PnP-Net and REPPnP on real data with different wrong matches number. PnP-Net on the left and REPPnP-LM on the right.} 
\label{fig:PnP_Net_REPPnP_LM_Wrong}
\end{figure}

To support our claims that PnP-Net does not overfit, we conducted additional experiments with various outlier generation mechanisms that were not used during training. In particular, we repeated the experiments with a constant number of mismatches (rather than a uniformly random number) and compared the leading competitor REPPnP-LM with PnP-Net. The results are presented in Fig. \ref{fig:PnP_Net_REPPnP_LM_Wrong} as a function of this number of mismatches. As expected, both algorithms fail when there are more than 3 outliers and the model is unidentifiable. With fewer outliers, PnP-Net consistently outperforms REPPnP-LM.



\section{Conclusion}
In this paper, we proposed a hybrid single-shot technique to recover the 6D pose of a camera. The architecture involves a deep neural network that recovers a coarse but robust estimate, followed by traditional model-based optimization techniques that refines the estimates. Both stages are differentiable and are optimized in an end to end manner. Our approach outperforms the state of the art by a significant margin in both synthetic and real data and shows the benefit of such hybrid approaches.

PnP-Net uses data associations that were collected by different algorithms as input. The common practice is to start by collecting objects from the image and matching them using a descriptor for each image detected and world object.
For example, detecting signs using a trained detection neural network and matching them to the world signs based on their shape and color.
Adding the detection and matching algorithms to the pipeline before PnP-Net and training both the detection network and our network in an end-to-end manner will probably improve both stages. 

Another interesting direction for future work is to train a neural network to output the likelihoods of each correspondence being wrong. These can be used as informative weights in the LM stage \cite{lm}. More generally, we believe the neural initialization, with and without the likelihoods, can be beneficial in other robust estimation problems. 


\section{Acknowledments}
The authors would like to thank Refael Vivanti for his help and advice. This work was partially supported by ISF grant 1339/15.

\bibliographystyle{plain}
\bibliography{references}

\end{document}